\documentclass[]{llncs}
\usepackage{url}
\usepackage{graphicx}
\usepackage{amsmath,amssymb} 
\usepackage{color}
\usepackage[width=122mm,left=12mm,paperwidth=146mm,height=193mm,top=12mm,paperheight=217mm]{geometry}
\usepackage[colorlinks,
            linkcolor=black,
            anchorcolor=black,
            citecolor=black,
            urlcolor=black
            ]{hyperref}
\begin{document}
\pagestyle{headings}
\mainmatter
\def\ECCV18SubNumber{1881}  

\title{Progressive DNN Compression: A Key to Achieve Ultra-High Weight Pruning and Quantization Rates using ADMM} 

\author{Shaokai Ye$^{1}$, Xiaoyu Feng$^2$, Tianyun Zhang$^{3}$, Xiaolong Ma$^4$, Sheng Lin$^4$, Zhengang Li$^4$, Kaidi Xu$^4$, Wujie Wen$^5$, Sijia Liu$^6$,  Jian Tang$^3$,  Makan Fardad$^3$, Xue Lin$^4$, Yongpan Liu$^2$ \& Yanzhi Wang$^{4*}$\\
1. Sensetime Research 2. Tsinghua University 3. Syracuse University \\ 4. Northeastern University  5. Florida International University \\ 6. MIT-IBM Watson AI Lab, IBM Research \\   
\texttt{* yanz.wang@northeastern.edu}\\
}
\institute{}


\maketitle

\begin{abstract}
Weight pruning and weight quantization are two important categories of DNN model compression. Prior work on these techniques are mainly based on heuristics. 
A recent work developed a systematic framework of DNN weight pruning using the advanced optimization technique ADMM (Alternating Direction Methods of Multipliers), achieving one of state-of-art in weight pruning results. 
In this work, we first extend such one-shot ADMM-based framework to guarantee solution feasibility and provide fast convergence rate, and generalize to weight quantization as well.
We have further developed a multi-step, progressive DNN weight pruning and quantization framework, with dual benefits of (i) achieving further weight pruning/quantization thanks to the special property of ADMM regularization, and (ii) reducing the search space within each step. Extensive experimental results demonstrate the superior performance compared with prior work.
   Some highlights: (i) we achieve 246$\times$, 36$\times$, and 8$\times$ weight pruning on LeNet-5, AlexNet, and ResNet-50 models, respectively, with (almost) zero accuracy loss; (ii) even a significant 61$\times$ weight pruning in AlexNet (ImageNet) results in only minor degradation in actual accuracy compared with prior work; (iii) we are among the first to derive notable weight pruning results for ResNet and MobileNet models; (iv) we derive the first lossless, fully binarized (for all layers) LeNet-5 for MNIST and VGG-16 for CIFAR-10; and (v) we derive the first fully binarized (for all layers) ResNet for ImageNet with reasonable accuracy loss.
Our models and sample codes are released in link \url{https://bit.ly/2TYx7Za}.
\end{abstract}

\section{Introduction}

Deep neural networks (DNNs) are both computationally and storage intensive \cite{krizhevsky2012imagenet,simonyan2014}. A number of prior work have focused on developing \emph{model compression} techniques for DNNs. These techniques, which are applied during the training phase of the DNN, aim to simultaneously reduce the model size (thus, the storage requirement) and accelerate the computation for inference -- all these to be achieved with minor classification accuracy (or prediction quality) loss. Indeed the  accuracy of a DNN inference engine after model compression is typically higher than that of a shallow neural network with no compression \cite{han2015learning,wen2016learning}. 
Two important categories of DNN model compression techniques are \emph{weight pruning} and \emph{weight quantization}.

An early work on weight pruning of DNNs was done by Han \emph{et al.} \cite{han2015learning}. It is an iterative heuristic method, achieving a 9$\times$ reduction in the number of weights of AlexNet model (for ImageNet dataset). This weight pruning method has been extended in \cite{dai2017,yang2016,guo2016dynamic,dong2017learning,wen2016learning,he2017channel} to either use more sophisticated algorithms to achieve a higher weight pruning rate, or to incorporate certain regularity or ``structures" in the weight pruning framework.
Weight quantization of DNNs has also been investigated in many recent work \cite{leng2017extremely,park2017weighted,zhou2017incremental,lin2016fixed,wu2016quantized,rastegari2016xnor,hubara2016binarized,courbariaux2015binaryconnect}. Both storage and computational requirements of DNNs have been greatly reduced with tolerable accuracy loss. Indeed, 
multiplication operations (which are costly) may be eliminated when using binary, ternary, or power-of-2 weight quantizations \cite{rastegari2016xnor,hubara2016binarized,courbariaux2015binaryconnect}.

To overcome the limitation of the highly heuristic nature in prior weight pruning work, a recent work \cite{zhang2018systematic} developed a systematic framework of DNN weight pruning using the advanced optimization technique ADMM (Alternating Direction Methods of Multipliers) \cite{boyd2011distributed,hong2016convergence}. Through the adoption of ADMM, the original weight pruning problem is decomposed into two subproblems, one effectively solved using stochastic gradient descent as original DNN training, while the other solved optimally and analytically via Euclidean projection \cite{zhang2018systematic}. This method achieves one of state-of-art in weight pruning results, 21$\times$ weight reduction in AlexNet and 71.2$\times$ in LeNet-5 without accuracy loss. However, the direct application of ADMM technique lacks the guarantee on solution feasibility (satisfying all constraints) due to the non-convex nature of objective function (loss function), while there is also margin of improvement for solution quality (in terms of pruning rate under the same accuracy).

In this work, we first make the following extensions on the one-shot ADMM-based weight pruning \cite{zhang2018systematic}: (i) develop an integrated framework of dynamic ADMM regularization and masked mapping and retraining steps, thereby guaranteeing solution feasibility and providing high solution quality; (ii) incorporate the multi-$\rho$ updating technique for faster (and better) ADMM convergence; and (iii) generalize to a unified framework applicable both the weight pruning and weight quantization. These extensions already provide higher performance than \cite{zhang2018systematic}.

Beyond the above extensions, we observe the opportunity of performing further weight pruning from the results of the one-shot ADMM-based weight pruning framework. This is due to the special property of $L_2$-based ADMM regularization process. Similar observation also applies to the weight quantization problem, and both suggest a \emph{progressive, multi-step model compression framework using ADMM}. In the progressive framework, the pruning/quantization results from the previous step serve as intermediate results and starting point for the subsequent step. It has an additional benefit of reducing the search space for weight pruning/quantization within each step. Detailed procedure and hyperparameter determination process have been carefully designed towards ultra-high weight pruning and quantization rates.

Extensive experimental results demonstrate that the proposed progressive framework consistently outperforms prior work. Some highlights: (i) we achieve 246$\times$, 36$\times$, and 8$\times$ weight pruning on LeNet-5, AlexNet, and ResNet-50 models, respectively, with (almost) zero accuracy loss; (ii) even a significant 61$\times$ weight pruning in AlexNet (ImageNet) results in only minor degradation in actual accuracy compared with prior work; (iii) we are among the first to derive notable weight pruning results for ResNet and MobileNet models; (iv) we derive the first lossless, fully binarized (for all layers) LeNet-5 for MNIST and VGG-16 model for CIFAR-10; and (v) we derive the first fully binarized (for all layers) ResNet model for ImageNet with reasonable accuracy loss.
Our models and sample codes are released in link \url{https://bit.ly/2TYx7Za}.


\section{Related Work}

\textbf{\emph{Weight pruning.}} An early work of weight pruning is \cite{han2015learning}. 
It uses a heuristic, iterative method to prune  weights of small magnitudes and retrain the DNN. 
It achieves 9$\times$ reduction in the number of weights on AlexNet for ImageNet dataset without accuracy degradation. 
However, this work achieves relatively low compression rate (2.7$\times$ for AlexNet) on CONV layers, which are the key computational part in state-of-the-art DNNs \cite{simonyan2015very,he2016deep}. 
Besides, indices are needed, at least one per weight, to index the relative location of the next weight. 
This method has been extended in two directions.
The first is to improve reduction in the number of weights by using more sophisticated heuristics, e.g., incorporating both weight pruning and growing \cite{guo2016dynamic}, using $L_1$ regularization \cite{wen2016learning}, or genetic algorithm \cite{dai2017nest}. 
The second is enhancing the actual implementation efficiency by deriving an effective tradeoff between accuracy and compression rate, e.g., the \emph{energy-aware pruning} \cite{yang2017designing}, and incorporating regularity in weight pruning, e.g., the \emph{channel pruning} \cite{he2017channel} and \emph{structured sparsity learning} \cite{wen2016learning} approaches. 

\textbf{\emph{Weight quantization.}} This method leverages the inherent redundancy in the number of bits for weight representation. 
Many of the  prior art work \cite{leng2017extremely,park2017weighted,zhou2017incremental,lin2016fixed,wu2016quantized,rastegari2016xnor,hubara2016binarized,courbariaux2015binaryconnect} are directed at quantization of weights to binary values, ternary values, or powers of 2 to facilitate hardware implementations, with acceptable accuracy loss. 
The state-of-the-art techniques \cite{courbariaux2015binaryconnect,leng2017extremely} adopt an iterative quantization and retraining framework, with some degree of randomness incorporated into the quantization step. 
This method results in less than 3\% accuracy loss on AlexNet for binary weight quantization \cite{leng2017extremely}.

\section{Overall Framework of Progressive DNN Model Compression}

\begin{figure} [!ht]
     \centering
     \includegraphics[width=0.8\columnwidth]{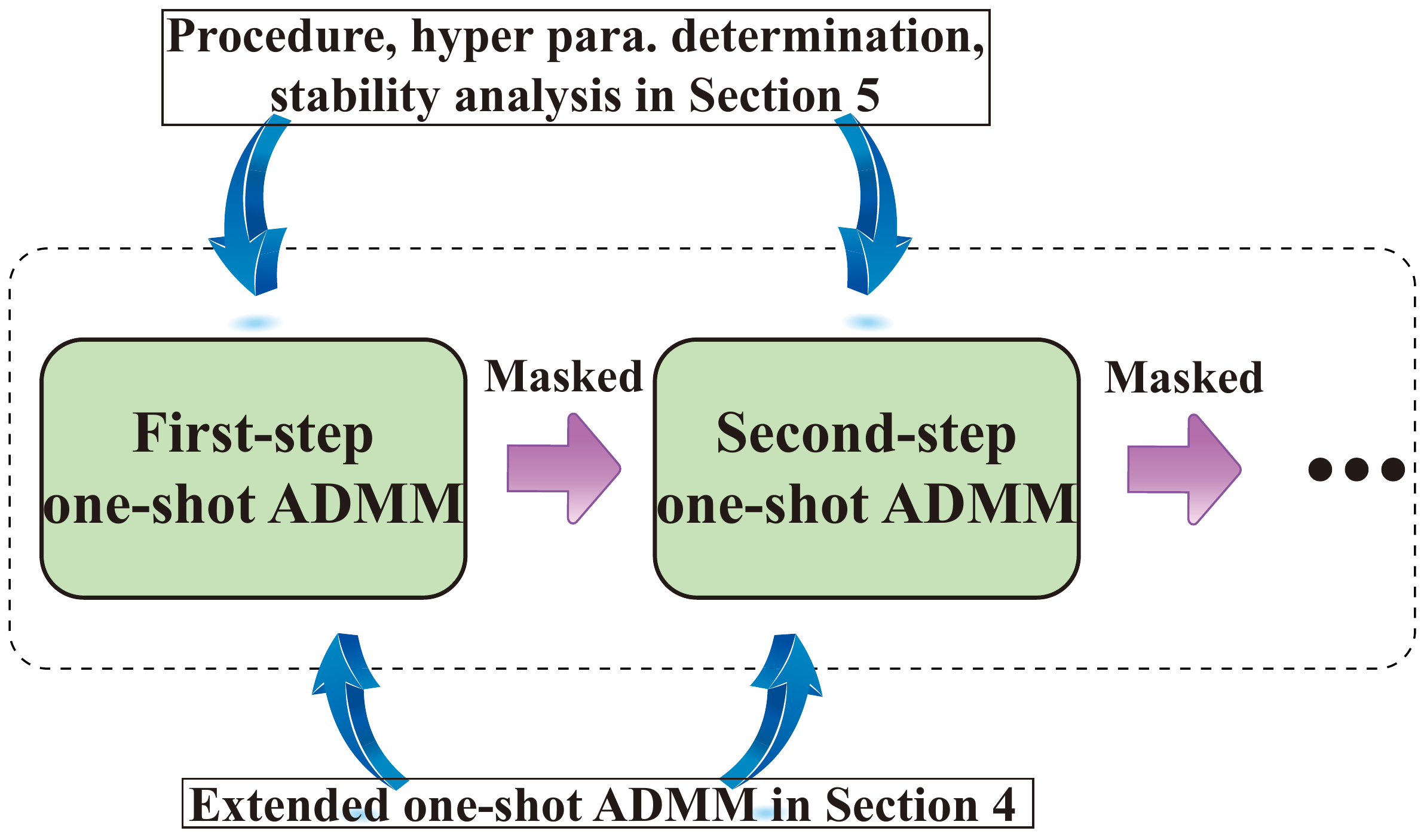}  
     \caption{Illustration of progressive DNN model compression.}
     \label{fig:general_step}
 \end{figure}

Figure \ref{fig:general_step} illustrates the proposed progressive DNN weight pruning and weight quantization framework. The one-shot ADMM-based weight pruning or quantization is performed multiple times, each as a step in the progressive framework. The pruning/quantization results from the previous step serve as intermediate results and starting point for the subsequent step. As discussed before, the reasons to develop a progressive model compression framework are twofold: (i) The fact that many weights are close to zero after ADMM regularization enables further weight pruning (such observation also applies to quantization); and (ii) the multi-step procedure reduces the search space for weight pruning/quantization within each step.  

Through extensive investigations, we conclude that a \underline{two-step progressive procedure} will be in general sufficient for weight pruning and quantization, in which each step requires approximately the same number of training epochs as original DNN training. Further increase in the number of steps or the number of epochs in each step will result in only marginal improvement in the overall solution quality (e.g., 0.1\%-0.2\% accuracy improvement).

The detailed description of the proposed progressive framework will be presented in Section \ref{sec:single} and Section \ref{sec:progressive}. Section \ref{sec:single} will present the proposed single-step, ADMM-based weight pruning and quantization framework, as an extension of \cite{zhang2018systematic} to guarantee solution feasibility and a generalization to weight quantization as well. Section \ref{sec:progressive} presents the motivation, detailed procedure, and hyperparameter determination of the proposed progressive model compression framework, along with illustration of why ``progressive" is the key to ultra-high compression rates.

\section{Single-Step, ADMM-based Weight Pruning and Quantization}\label{sec:single}

\subsection{Optimization Problem Formulation}

Consider an $N$-layer DNN with both CONV and FC layers. The weights and biases of the $i$-th layer are respectively denoted by ${\bf{W}}_{i}$ and ${\bf{b}}_{i}$, and the loss function associated with the DNN is denoted by $f \big( \{{\bf{W}}_{i}\}_{i=1}^N, \{{\bf{b}}_{i} \}_{i=1}^N \big)$; see \cite{zhang2018systematic}. In this paper, $\{{\bf{W}}_{i}\}_{i=1}^N$ and $\{{\bf{b}}_{i} \}_{i=1}^N$ respectively characterize the collection of weights and biases from layer $1$ to layer $N$. Then DNN weight pruning or weight quantization is formulated as the following optimization problem:
\begin{equation}
\label{opt0}
\begin{aligned}
& \underset{ \{{\bf{W}}_{i}\},\{{\bf{b}}_{i} \}}{\text{minimize}}
& & f \big( \{{\bf{W}}_{i}\}_{i=1}^N, \{{\bf{b}}_{i} \}_{i=1}^N \big),
\\ & \text{subject to}
& & {\bf{W}}_{i}\in {\mathcal{S}}_{i}, \; i = 1, \ldots, N,
\end{aligned}
\end{equation}

For \underline{weight pruning}, the constraint set is ${\mathcal{S}}_{i}=\big\{{{\bf{W}}_{i}\big|\text{card}(\text{supp}({\bf{W}}_{i}))\le \alpha_i\big\}}$, where `card' refers to cardinality and `supp' refers to the support set. Elements in ${\mathcal{S}}_{i}$ are ${\bf{W}}_{i}$ solutions, satisfying that the number of non-zero elements in ${\bf{W}}_{i}$ is limited by $\alpha_i$ for layer $i$. These $\alpha_i$ values are hyperparameters, with determination heuristic in Section \ref{sec:progressive}. Besides the general, non-structured weight pruning scenario, the constraint set can be extended to \underline{incorporate specific ``structures"} corresponding to structured pruning techniques such as filter pruning, channel pruning, column pruning, etc., with detailed discussions in \cite{zhang2018adam}. The appropriate structured pruning will facilitate high-parallelism implementations in hardware\footnote{The default weight pruning in this paper is the general, non-structured pruning. However, the proposed framework is also applicable to structured weight pruning, with results in supplementary materials.}.

For \underline{weight quantization}, elements in the constraint set ${\mathcal{S}}_{i}$ are ${\bf{W}}_{i}$ solutions, in which elements in ${\bf{W}}_{i}$ assume one of  $q_{i,1},q_{i,2},...,q_{i,M_i}$ values, where $M_i$ denotes the number of these fixed values. Here, the $q_{i,j}$ values are \emph{quantization levels} of weights of layer $i$ in increasing order, and we focus on \emph{equal-distance quantization} (the same distance between adjacent quantization levels) to facilitate hardware implementations. 
For the \underline{combination of weight pruning and quantization} for DNNs, it is common practice to perform weight pruning first, and then quantization on the remaining, non-zero weights.

\subsection{A Unified Solution Framework using ADMM}

In problem (\ref{opt0}) the constraint is combinatorial. 
As a result, this problem cannot be solved directly by stochastic gradient descent methods like original DNN training.
However, the form of the combinatorial constraints on ${\bf{W}}_{i}$ is compatible with ADMM which is recently shown to be an effective method to deal with such clustering-like constraints \cite{hong2016convergence,liu2018zeroth}

Despite such compatibility, there is still challenge in the direct application of ADMM due to the non-convexity in objective function. To overcome this challenge, we extend over \cite{zhang2018systematic} and develop a systematic framework of dynamic ADMM regularization and masked mapping and retraining steps. We can guarantee solution feasibility (satisfying all constraints) and provide high solution quality through this integration. This framework is unified and applies to both weight pruning and weight quantization, and will act as one step in the progressive DNN weight pruning/quantization framework. 

\begin{figure} [!ht]
     \centering
     \includegraphics[width=0.78\columnwidth]{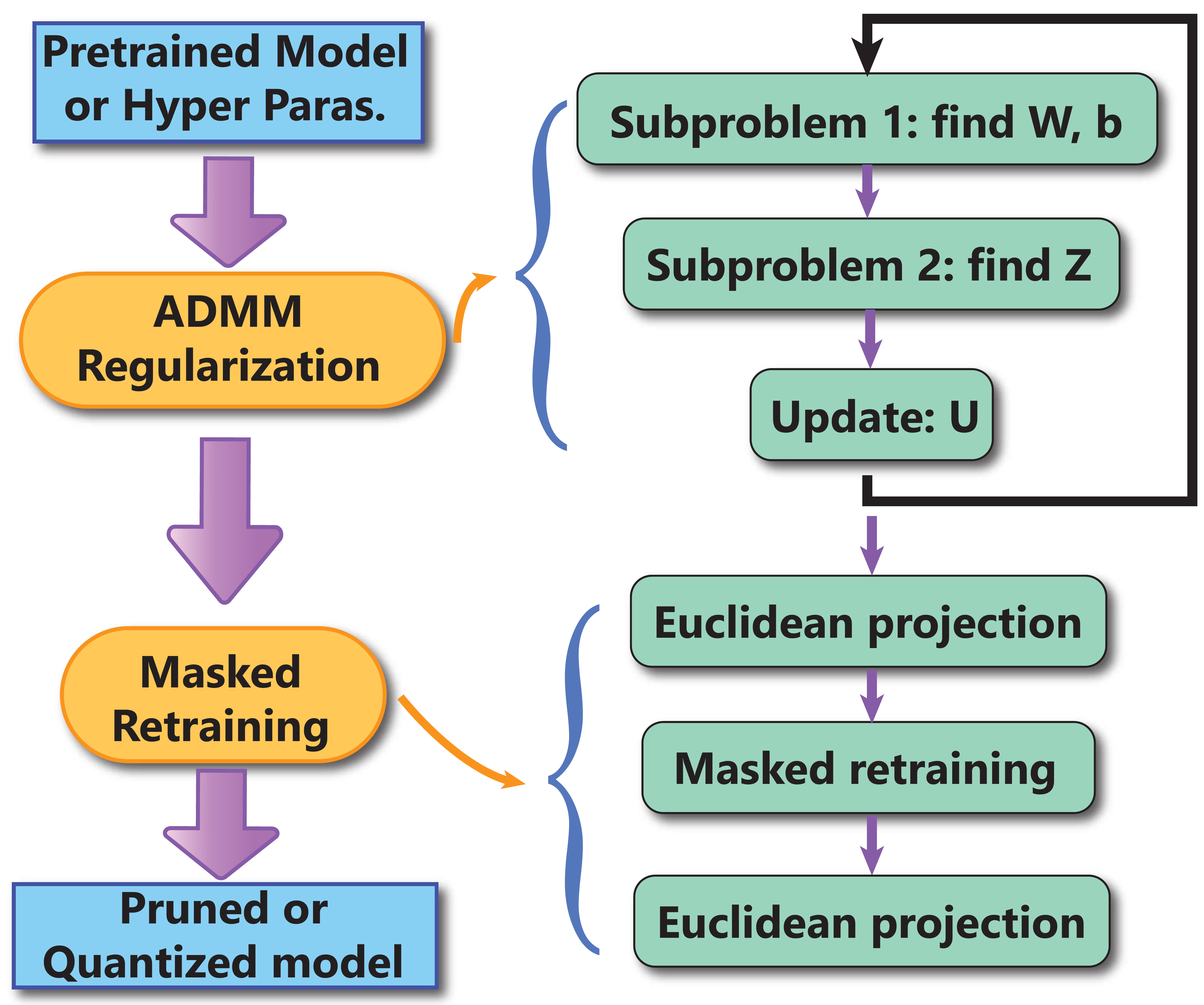}    
     \caption{Procedure of one-shot ADMM-based weight pruning/quantization}
     \label{fig:admm_general}
 \end{figure}


\emph{\textbf{ADMM Regularization Step}}: Corresponding to every set ${\mathcal{S}}_{i}$, $i = 1, \ldots, N$ we define the indicator function
$g_{i}({\bf{W}}_{i})=
\begin{cases}
 0 & \text { if } {\bf{W}}_{i}\in {\mathcal{S}}_{i}, \\ 
 +\infty & \text { otherwise.}
\end{cases}$
Furthermore, we incorporate auxilliary variables ${\bf{Z}}_{i}$, $i = 1, \ldots, N$.
The original problem (\ref{opt0}) is then equivalent to 
\begin{equation}
\label{admm_form}
\begin{aligned}
& \underset{ \{{\bf{W}}_{i}\},\{{\bf{b}}_{i} \}}{\text{minimize}}
& & f \big( \{{\bf{W}}_{i} \}_{i=1}^N, \{{\bf{b}}_{i} \}_{i=1}^N \big)+\sum_{i=1}^{N} g_{i}({\bf{Z}}_{i}),
\\ & \text{subject to}
& & {\bf{W}}_{i}={\bf{Z}}_{i}, \; i = 1, \ldots, N.
\end{aligned}
\end{equation}

Through formation of the augmented Lagrangian \cite{boyd2011distributed}, the ADMM regularization decomposes problem (\ref{admm_form}) into two subproblems, and solves them iteratively until convergence\footnote{The details of ADMM are presented in \cite{boyd2011distributed,zhang2018systematic}. We omit the details due to space limitation.}. The first subproblem is
\begin{equation}
\label{subproblem_1}
 \underset{ \{{\bf{W}}_{i}\},\{{\bf{b}}_{i} \}}{\text{minimize}}
\ \ \ f \big( \{{\bf{W}}_{i} \}_{i=1}^N, \{{\bf{b}}_{i} \}_{i=1}^N \big)+\sum_{i=1}^{N} \frac{\rho_{i}}{2}  \| {\bf{W}}_{i}-{\bf{Z}}_{i}^{k}+{\bf{U}}_{i}^{k} \|_{F}^{2}, \\
\end{equation}
where ${\bf{U}}_{i}^{k}:={\bf{U}}_{i}^{k-1}+{\bf{W}}_{i}^{k}-{\bf{Z}}_{i}^{k}$.
The first term in the objective function of (\ref{subproblem_1}) is the differentiable loss function of the DNN, and the second term is a quadratic regularization term of the ${\bf{W}}_{i}$'s, which is differentiable and convex. As a result (\ref{subproblem_1}) can be solved by stochastic gradient descent as original DNN training. Although we cannot guarantee the global optimality, it is due to the non-convexity of the DNN loss function rather than the quadratic term enrolled by our method. Please note that this first subproblem maintains the \underline{same form for weight pruning and quantization problems}.

On the other hand, the second subproblem is given by
\begin{equation}
 \underset{ \{{\bf{Z}}_{i} \}}{\text{minimize}}
\ \ \ \sum_{i=1}^{N} g_{i}({\bf{Z}}_{i})+\sum_{i=1}^{N} \frac{\rho_{i}}{2} \| {\bf{W}}_{i}^{k+1}-{\bf{Z}}_{i}+{\bf{U}}_{i}^{k} \|_{F}^{2}. \\
\end{equation}
Note that $g_{i}(\cdot)$ is the indicator function of ${\mathcal{S}}_{i}$, thus this subproblem can be solved analytically and optimally \cite{boyd2011distributed}. For $i = 1, \ldots, N$, the optimal solution is the Euclidean projection of ${\bf{W}}_{i}^{k+1}+{\bf{U}}_{i}^{k}$ onto ${\mathcal{S}}_{i}$. For \underline{weight pruning}, we can prove that the Euclidean projection results in keeping $\alpha_i$ elements in ${\bf{W}}_{i}^{k+1}+{\bf{U}}_{i}^{k}$ with the largest magnitudes and setting the remaining weights to zeros. For \underline{weight quantization}, we can prove that the Euclidean projection results in mapping every element of ${\bf{W}}_{i}^{k+1}+{\bf{U}}_{i}^{k}$ to the quantization level closest to that element.

After both subproblems solved, we update the dual variables ${\bf{U}}_{i}$'s according to the ADMM rule \cite{boyd2011distributed} and thereby complete one iteration in ADMM regularization.

\emph{\textbf{Increasing $\rho$ in ADMM regularization}}: The $\rho_{i}$ values are the most critical hyperparameter in ADMM regularization. We start from smaller $\rho_{i}$ values, say $\rho_{1} = \dots = \rho_{N} = 1.5\times 10^{-3}$, and gradually increase with ADMM iterations. This coincides with the theory of ADMM convergence \cite{hong2016convergence,liu2018zeroth}. It in general takes 8 - 12 ADMM iterations for convergence (more iterations to converge for weight pruning and fewer for weight quantization), corresponding to 100 - 150 epochs in PyTorch. This convergence rate is comparable with the original DNN training.

\emph{\textbf{Masked mapping and retraining}}: After ADMM regularization, we obtain intermediate ${\bf{W}}_{i}$ solutions. The subsequent step of masked mapping and retraining will guarantee the solution feasibility and improve solution quality. For \underline{weight pruning}, the procedure is more straightforward. We first perform the said Euclidean projection (mapping) to guarantee that pruning constraints are satisfied. Next, we mask the zero weights and retrain the DNN with non-zero weights using training sets (while keeping the masked weights 0). In this way test accuracy (solution quality) can be (partially) restored, and solution feasibility (constraints) will be maintained.

For \underline{weight quantization}, the procedure is more complicated. The reason is that the retraining process will affect the quantization results, thereby solution feasibility. To deal with this issue, we first perform Euclidean projection (mapping) of weights that are close enough (defined by a threshold value $\epsilon$) to nearby quantization levels. Then we perform retraining on the remaining, unquantized weights (with quantized weights fixed) for accuracy improvement. Finally we perform Euclidean mapping on the remaining weights as well. In this way the solution feasibility will be guaranteed.

\subsection{Explanation of Effectiveness in the Deep Learning Context}\label{sec:single-illu}

The proposed solution framework is different from the conventional utilization of ADMM, i.e., to accelerate the convergence of an originally convex problem \cite{boyd2011distributed,hong2017linear}. Rather, we integrate the ADMM framework with stochastic gradient descent. Aside from recent mathematical optimization results \cite{hong2016convergence,liu2018zeroth} illustrating the advantage of ADMM with combinatorial constraints, the advantage of the proposed solution framework can be explained in the deep learning context as described below.

The proposed solution (\ref{subproblem_1}) can be understood as a smart, dynamic $L_2$ regularization method, in which the regularization target $ {\bf{Z}}_{i}^{k}-{\bf{U}}_{i}^{k}$ will change judiciously and analytically in each iteration. On the other hand, conventional regularization methods (based on $L_1$, $L_2$ norms or their combinations) use a fixed regularization target, and the penalty is applied on all the weights. This will inevitably cause accuracy degradation. More illustrations of the ADMM-based dynamic regularization vs. conventional, fixed regularization will be provided in Section \ref{sec:progressive-illu}.

\section{Progressive DNN Model Compression Framework: Detailed Procedure}\label{sec:progressive}

\subsection{Motivation}
During the implementation of the one-shot weight pruning framework described in Section \ref{sec:single}, we observe that there are a number of unpruned weights with values very close to zero. The reason is the $L_2$ regularization nature in ADMM regularization step, which tends to generate very small, non-zero weight values even when they are not pruned. As the remaining number of non-zero weights is already significantly reduced during weight pruning, simply mapping these small-value weights to zero will result in accuracy degradation. On the other hand, this motivates us to perform weight pruning (and quantization) in a multi-step, progressive manner. For weight pruning, the weights that have been pruned in the previous step will be masked and only the remaining, non-zero weights will be considered in the subsequent step. For weight quantization, we perform quantization on the weights in a subset of layers, fix these quantization results, and quantize the remaining layers in the subsequent step.

A second motivation of the progressive framework is to reduce the search space for weight pruning/quantization within each step. After all, weight pruning and quantization problems are essentially combinatorial optimizations. Although recently demonstrated to generate superior results on this type of problems \cite{hong2016convergence,liu2018zeroth}, ADMM-based solution still has a superlinear increase of computational complexity as a function of solution space. As a result, the complexity becomes very high with ultra-high compression rates (i.e., very large search space) beyond what can be achieved in prior work. The progressive framework, on the other hand, can mitigate this limitation and reduce the total training time (to 2$\times$ or slightly higher than training time of the original DNN).

\subsection{Detailed Procedure and Hyperparameter Determination}

Through extensive investigations, we conclude that a \underline{two-step progressive procedure} will be in general sufficient for weight pruning and quantization, in which each step requires approximately the same number of training epochs as original DNN training. We have conducted experiments on CIFAR-10 and ImageNet benchmarks (AlexNet and ResNet-18 models) on the relative accuracy of two-step procedure vs. three-step procedure, in which each step uses 120 epochs for training in PyTorch. The results show that three-step procedure only possesses marginal improvement in the overall solution quality, i.e., accuracy improvement no greater than 0.2\%. This makes the additional training time not entirely worthwhile.

\emph{\textbf{Hyperparameter Determination and Sensitivity Analysis}}: A very critical question is how to determine the hyperparameters, in a highly efficient and reliable manner. This problem is challenging for weight pruning, because we need to determine both the target overall pruning rate and specific pruning rate for each layer, both required in the ADMM-based solution. For quantization it becomes relatively straightforward, as the target number of quantization bits is typically pre-specified (binary, ternary, 2-bit, etc.) and the same number of quantization bits for all layers is in general preferred in hardware. The objective is to minimize accuracy loss. As a result, the two-step procedure of weight quantization can be performed as follows: the first step performs quantization on all the weights except for the first and last layers, while the second step performs quantization on these two layers. This is because quantization on these two layers has more significant impact on the overall accuracy.

Let us focus again on the hyperparameter determination heuristic for weight pruning problems. Experiments demonstrate that at least 2$\times$ to 3$\times$ improvement in overall pruning rate can be achieved compared with the prior work \cite{han2015learning}, under the same accuracy or without accuracy loss. Again at least 50\% improvement in pruning rate can be achieved compared with the prior work of one-shot ADMM-based weight pruning \cite{zhang2018systematic}. As a result, a simple but effective \underline{hyperparameter determination method} is as follows: We set the target overall pruning rate in the first ADMM-based weight pruning step to be around 1.5$\times$ compared with what can be achieved (without accuracy loss) in prior work \cite{han2015learning}, or to be slightly lower than the final result in \cite{zhang2018systematic}. The target overall pruning rate in the second step will be doubled compared with the first step, or even further increased if there is still margin of improvement. The per-layer pruning rate will be inherited from the results in prior work and increased proportionally. According to our experiments, the above heuristic will generate consistently higher pruning rates than prior work without accuracy loss.

We have further conducted two experiments to demonstrate the stability of hyperparameter (per-layer pruning rates) selection. Detailed experimental setup and results are provided in supplementary materials. The general conclusions are: (i) certain degree of variations in the per-layer pruning rates will have only minor impact on the overall accuracy under the ADMM-based framework; (ii) for very deep DNNs such as ResNet-50, uniform pruning rates for all layers will result in a reasonably good overall pruning results. These results demonstrate the robustness of the hyperparameter determination process.

Although the above discussions are based on the general, non-structured weight pruning, the above hyperparameter determination is also applicable to structured pruning.

\subsection{Discussions and Illustration of Effectiveness through Weight Pruning}\label{sec:progressive-illu}

\begin{figure}[ht]
\centering
\includegraphics[width=0.83\linewidth]{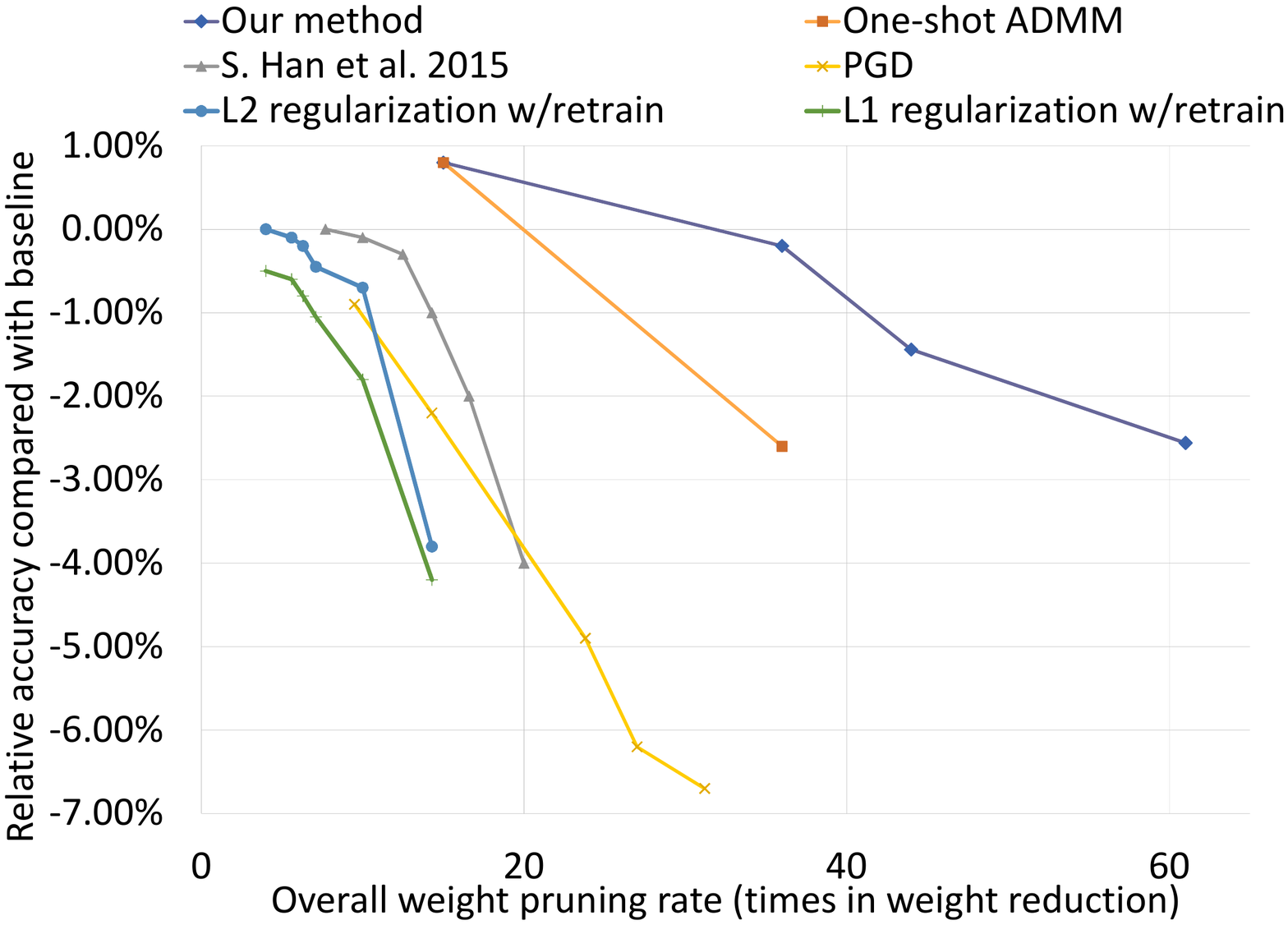}
\caption{Relative top-5 accuracy compared with baseline for different pruning methods on AlexNet for ImageNet data set.}
\label{fig:example1}
\end{figure}

\begin{figure}[ht]
\centering
\includegraphics[width=0.83\linewidth]{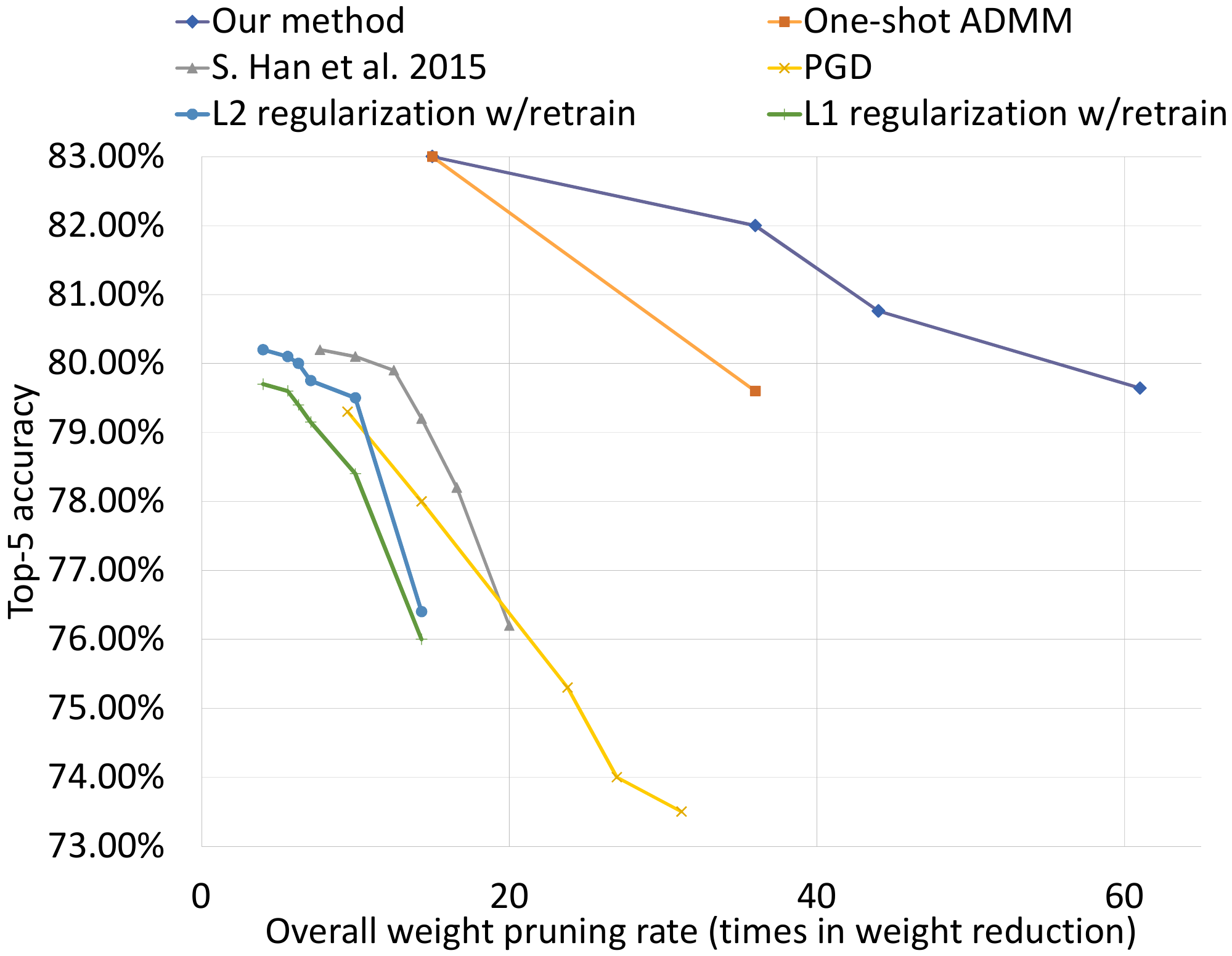}
\caption{Absolute top-5 accuracy for different pruning methods on AlexNet for ImageNet data set.}
\label{fig:example2}
\end{figure}

Using AlexNet model on ImageNet data set as an example, Figure \ref{fig:example1} demonstrates the Top-5 accuracy loss vs. overall pruning rates using various methods, including our proposed progressive framework, our enhanced one-shot ADMM-based pruning, iterative pruning and retraining reported in \cite{han2015learning}, $L_1$ and $L_2$ fixed regularizations and projected gradient descent (PGD). Figure \ref{fig:example2} demonstrates the absolute Top-5 accuracy. Please note that we use a baseline AlexNet model with 60.0\% Top-1 accuracy and 82.2\% Top-5 accuracy, both higher than prior work such as \cite{han2015learning,zhang2018systematic} (57.2\% Top-1 and 80.2\% Top-5). This is to reflect the recent advances in DNN training in PyTorch. As a result, our definition of accuracy loss (or lossless) is compared with respect to the enhanced accuracy. In other words, we aim to surpass the prior methods in both absolute accuracy and relative accuracy loss values.

We can clearly observe the performance ranking of these techniques. The proposed progressive framework outperforms all other methods. The second is one-shot ADMM-based pruning. The third is iterative pruning and retraining heuristic. And the last is fixed regularizations and PGD. We know from Section \ref{sec:single-illu} that fixed regularizations and PGD suffer from penalizing all weights even if they are not pruned, thereby resulting in notable accuracy degradation. Then how to explain the performance gap among the other techniques?

\begin{figure} [!ht]
     \centering
     \includegraphics[width=0.9\columnwidth]{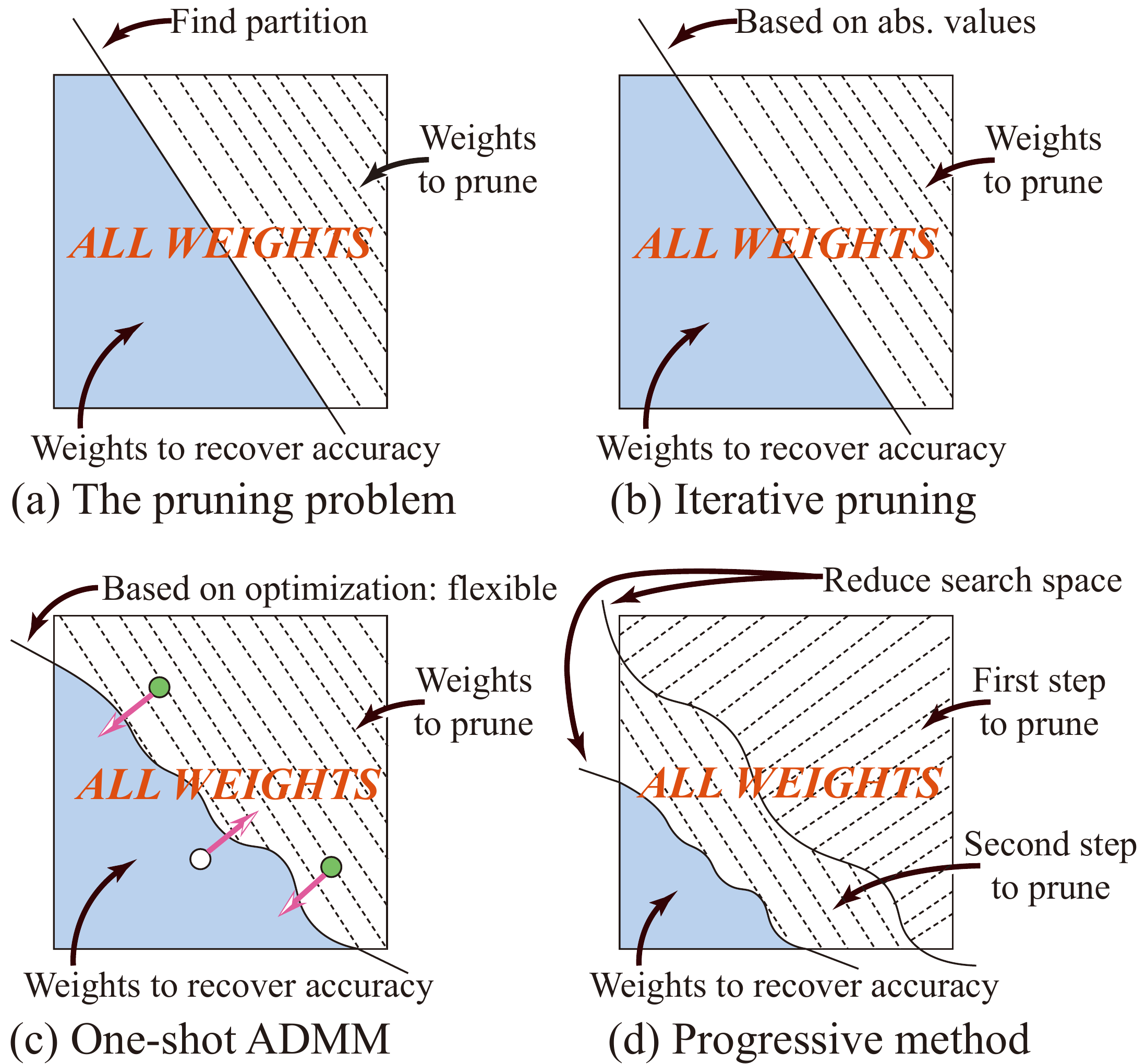}    
     \caption{Illustration of effectiveness of the one-shot ADMM-based weight pruning and the progressive method.}
     \label{fig:diff_prune}
 \end{figure}

To answer this question, we use Figure \ref{fig:diff_prune} as an illustration. The weight pruning problem can be understood as a \emph{partitioning problem}, in which weights will be partitioned into two parts, one part all mapped to zero, while the other part utilized to restore accuracy. The straightforward iterative pruning method performs partitioning based only on the absolute values of the weights, smaller ones mapped to zero. The ADMM-based weight pruning method, on the other hand, allows partitioning using effective mathematical optimization methods, thereby achieving higher pruning rates without accuracy loss. Then new challenge exists on the high complexity in deriving such partitioning when the pruning rates become ultra-high, and this challenge can be effectively mitigated using the progressive method by reducing the search space within each step.

\section{Experimental Results}
In this section, we evaluate the proposed progressive DNN model compression framework comprehensively, based on ImageNet ILSVRC-2012, CIFAR-10, and MNIST data sets, using AlexNet \cite{krizhevsky2012imagenet}, VGGNet \cite{simonyan2014}, ResNet-18/ResNet-50 \cite{he2016deep}, MobileNet V2 \cite{sandler2018mobilenetv2}, and LeNet-5 DNN models, and comparing with various prior methods including single-shot ADMM. Our implementations are based on PyTorch, and the baseline accuracies are in many cases higher than those utilized in prior work, such as AlexNet and ResNet-50 for ImageNet, VGGNet and MobileNet V2 for CIFAR-10, etc. We conduct a fair comparison because \underline{we focus on the relative accuracy} with our baseline instead of the absolute accuracy (which will of course outperform prior work).

Thanks to the compatibility of the proposed framework with DNN training, directly training a DNN model using the proposed framework has the same result as using a prior pre-trained DNN model. When a pre-trained DNN model is utilized, we limit the number of epochs in both steps in the progressive framework to be 120, similar to the original DNN training in PyTorch and much lower than the iterative pruning heuristic \cite{han2015learning}. We use the hyperparameter determination procedure discussed in Section \ref{sec:progressive-illu}. The training and model compression are performed in PyTorch using NVIDIA 1080Ti, 2080, and Tesla P100 GPUs.

Due to space limitation, in this section we only present results on the general, non-structured weight pruning and sample results on binary quantizations. More comprehensive results on structured weight pruning, combination of weight pruning and quantization, and convergence analysis are provided in the supplementary materials.

\subsection{Experimental Results on Weight Pruning}

\subsubsection{Results on ImageNet Dataset}

\begin{table}
\centering
\caption{Overall weight pruning rate comparisons on AlexNet model for ImageNet data set.}\label{table:AlexNet1}
\begin{tabular}{p{2cm}p{1.5cm}p{1.7cm}p{1.4cm}}
\hline
Method &  Top-5 accuracy & Relative accuracy loss  & Overall prun. rate  \\ \hline
SVD \cite{denton2014exploiting} & $79.4$\% & $+0.9$\% & 5.1$\times$ \\ \hline
Iter. prun. \cite{han2015learning} & $80.3$\%  & $-0.1$\% & 9.1$\times$ \\ \hline
NeST \cite{dai2017} & $80.3\%$ & $-0.1$\% & 15.7$\times$ \\ \hline
Dyn. surg. \cite{guo2016dynamic} & $80.0\%$ & $+0.2$\% & 17.7$\times$ \\ \hline
One-shot ADMM \cite{zhang2018systematic} & $80.2\%$ & $-0.0$\% & 17.7$\times$ \\ \hline
\bf{Our one-shot} & $83.0\%$ & $-0.8$\% & 15$\times$ \\ \hline
\bf{Our one-shot} & $79.6\%$ & $+2.6$\% & 36$\times$ \\ \hline
\bf{Our method} & $82.0\%$ & $+0.2$\% & 36$\times$ \\ \hline
\bf{Our method} & $80.8\%$ & $+1.4$\% & 44$\times$ \\ \hline
\bf{Our method} & $79.7\%$ & $+2.5$\% & 61$\times$ \\ \hline
\end{tabular}
\end{table}

\begin{table}
\centering
\caption{Convolutional layers weight pruning rate comparisons on the AlexNet model for ImageNet data set.}\label{table:AlexNet2}
\begin{tabular}{p{2.5cm}p{1.5cm}p{1.7cm}p{1.5cm}}
\hline
Method &  Top-5 accuracy & Relative accuracy loss  & Conv. prun. rate  \\ \hline
Iter. prun. \cite{han2015learning} & $80.3$\%  & $-0.1$\% & 2.7$\times$ \\ \hline
Dyn. surg. \cite{guo2016dynamic} & $80.0\%$ & $+0.2$\% & 3.1$\times$ \\ \hline
NeST \cite{dai2017} & $80.3\%$ & $-0.1$\% & 3.2$\times$ \\ \hline
Fine-grained \cite{mao2017exploring} & $80.4\%$ & $-0.2$\% & 4.2$\times$ \\ \hline
$L_1$ method \cite{wen2016learning} & $80.5\%$ & $-0.3$\% & 5.0$\times$ \\ \hline
\bf{Our method} & $82.4\%$ & $-0.2$\% & 8.6$\times$ \\ \hline
\bf{Our method} & $81.9\%$ & $+0.3$\% & 11.2$\times$ \\ \hline
\end{tabular}
\end{table}

\emph{\textbf{AlexNet Results}}: Table \ref{table:AlexNet1} compares the overall pruning rates of the whole AlexNet model (CONV and FC layers) vs. accuracy, between the proposed progressive framework and various prior methods. It can be clearly observed that the proposed framework outperforms prior methods, including the one-shot ADMM method \cite{zhang2018systematic}. With almost no Top-5 accuracy loss (note of our high baseline accuracy), we achieve 36$\times$ overall pruning rate. We achieve a notable 61$\times$ weight reduction with 79.7\% Top-5 accuracy, just slightly below the baseline accuracy in prior work. We can clearly observe the advantage over one-shot ADMM method. With the same accuracy, the progressive framework achieves 61$\times$ weight reduction while our extended one-shot method achieves ``only" 36$\times$. This 36$\times$ in one-shot method has been derived using the same number of total training epochs as the progressive framework.

Table \ref{table:AlexNet2} compares the pruning rates on the CONV layers vs. Top-5 accuracy, since the CONV layers are the most computationally intensive in state-of-art DNNs. We achieve 8.6$\times$ pruning in CONV layers with even slight accuracy enhancement, and 11.2$\times$ pruning with minor accuracy loss, consistently outperforming prior work in CONV layer weight pruning.

\emph{\textbf{VGG-16 Results}}: We conduct experiments on VGG-16 for ImageNet data set, with results similar to AlexNet. We achieve 34$\times$ overall weight reduction without accuracy loss, which is higher than 13$\times$ using iterative pruning \cite{han2015learning}, 15$\times$ in \cite{yu2017compressing} or 19.9$\times$ using our extended one-shot ADMM (no corresponding results reported in \cite{zhang2018systematic}). Detailed table is omitted due to space limitation.

\begin{table}[ht]
\centering
\caption{Comparisons of overall weight pruning results on ResNet-50 for ImageNet data set.}\label{table:ResNet-50}
\begin{tabular}{p{3cm}p{2.5cm}p{1.8cm}}
\hline
Method & Top-5 Acc. Loss  & Pruning rate \\ 
\hline
Uncompressed & 0.0\%  & 1$\times$ \\ \hline
 {Fine-grained \cite{mao2017exploring}}  & 0.1\%  & 2.6$\times$ \\ 
\hline
\bf{Our one-shot}  & 0.0\%  & 4.5$\times$ \\ \hline
\bf{Our method}  & 0.0\%  & 8$\times$ \\ \hline
\bf{Our method}  & 0.7\%  & 17.4$\times$ \\ \hline
\end{tabular}
\end{table}

\emph{\textbf{ResNet-18/ResNet-50 Results}}: We conduct experiments on ResNet-18 and ResNet-50 models for ImageNet data set. As there is lack of effective pruning results before, we conduct uniform weight pruning (the same pruning rate for all CONV and FC layers) to show the effectiveness with less optimized individual-layer pruning rates. The results are shown in Table \ref{table:ResNet-50}. We achieve 8$\times$ overall pruning rate (also 8$\times$ pruning rate on CONV layers) on ResNet-50, without accuracy loss. We also achieve 6$\times$ overall pruning rate (also 6$\times$ pruning rate on CONV layers) on ResNet-18, without accuracy loss. These results clearly outperform the prior work which has limited overall pruning rate, which also did not mention CONV layer rate. It also outperforms our one-shot ADMM-based method, which achieves 4.5$\times$ uniform weight pruning on all layers (CONV and FC) on ResNet-50.

\subsubsection{Results on CIFAR-10 Dataset}

\emph{\textbf{VGG-16 Results}}: We conduct experiments on VGG-16 results using the CIFAR-10 data set. The baseline accuracy is 93.7\%, which is higher than those in prior work, e.g., 90.2\% in \cite{qin2018demystifying} or 84.8\% in \cite{cheng2018differentiable}. We only present our results due to lack of prior work for fair comparison. We achieve 11.5$\times$ overall weight pruning without accuracy loss, or 40.3$\times$ with accuracy loss of 0.8\%.

\emph{\textbf{MobileNet V2 Results}}: We conduct experiments on MobileNet V2 results using the CIFAR-10 data set. The baseline accuracy is as high as 95.07\% due to the adoption of mixup technique. We present our results in Table \ref{table:MobileNet} due to lack of prior work for fair comparison. We achieve 5$\times$ weight pruning with almost no accuracy loss, starting from the high-accuracy baseline. We achieve 10$\times$ weight pruning (which is highly challenging for MobileNet) with only 1.3\% accuracy loss.

\begin{table}[ht]
\centering
\caption{Our weight pruning results on MobileNet V2 for CIFAR-10 data set.}\label{table:MobileNet}
\begin{tabular}{p{3cm}p{2.5cm}p{1.8cm}}
\hline
Method & Accuracy  & Pruning rate \\ 
\hline
Uncompressed & 95.07\%  & 1$\times$ \\ \hline
\bf{Our method}  & 95.49\%  & 3.3$\times$ \\ 
\hline
\bf{Our method}  & 94.90\%  & 5$\times$ \\ \hline
\bf{Our method}  & 94.70\%  & 6.7$\times$ \\ \hline
\bf{Our method}  & 93.75\%  & 10$\times$ \\ \hline
\end{tabular}
\end{table}

\subsubsection{Results on MNIST Dataset}

Table \ref{table:LeNet-5} demonstrates the comparison results on LeNet-5 model using MNIST data set. Through the progressive framework, we achieve an unprecedented 246$\times$ overall weight reduction with almost no accuracy loss. It clearly outperforms one-shot ADMM (71.2$\times$ using prior one-shot ADMM \cite{zhang2018systematic} and 85$\times$ using our extended one-shot ADMM) and other prior methods. Please note that our extended one-shot ADMM-based method also slightly outperforms the prior counterpart \cite{zhang2018systematic}.

\begin{table}[ht]
\centering
\caption{Comparisons of overall weight pruning results on LeNet-5 for MNIST data set.}\label{table:LeNet-5}
\begin{tabular}{p{3.5cm}p{1.5cm}p{1.8cm}}
\hline
Method & Accuracy &  Pruning rate \\ 
\hline
Uncompressed & 99.2\% &  1$\times$ \\ \hline
 {Network Pruning \cite{han2015learning}}  & 99.2\% &  12.5$\times$ \\ \hline
 {One-shot ADMM \cite{zhang2018systematic}}  & 99.2\% &  71.2$\times$ \\ \hline
 {Optimal Brain Surg. \cite{dong2017learning}}  & 98.3\% &  111$\times$ \\ \hline
 \bf{Our one-shot}  & 99.2\% &  85$\times$ \\ \hline
\bf{Our method}  & 99.2\% &  200$\times$ \\ \hline 
\bf{Our method}  & 99.0\% &  246$\times$ \\ \hline
\end{tabular}
\end{table}

\subsection{Sample Results on Weight Quantization}

\emph{\textbf{Binary Weight Quantization Results on LeNet-5}}: To the extent of authors' knowledge, we achieve \underline{the first lossless, fully binarized LeNet-5 model} in which weights in all layers are binarized. The accuracy is still 99.21\%, lossless compared with baseline. We do not list the comparison results due to limited space, but claim that our method already achieves the highest possible accuracy. We claim that becoming lossless is challenging even for MNIST. For example, recent work \cite{cheng2018differentiable} results in 2.3\% accuracy degradation on MNIST for full binarization, with baseline accuracy 98.66\%.

\emph{\textbf{Weight Quantization on CIFAR-10}}: We also achieve the first lossless, fully binarized VGG-16 for CIFAR-10, in which weights in all layers (including the first and the last) are binarized. The accuracy is 93.53\%. We would like to point out that fully ternarized quantization results in 93.66\% accuracy. Table \ref{table:VGG-quan} shows our results and comparisons.

\begin{table}[ht]
\centering
\caption{Comparisons of fully binary (ternary) weight quantization results on VGG-16 for CIFAR-10 data set.}\label{table:VGG-quan}
\begin{tabular}{p{2.5cm}p{1.5cm}p{2.5cm}}
\hline
Method & Accuracy &  Num. of bits \\ 
\hline
Baseline of \cite{cheng2018differentiable} & 84.80\% &  32 \\ \hline
 8-bit \cite{cheng2018differentiable} & 84.07\% &  8 \\ \hline
 Binary \cite{cheng2018differentiable}  & 81.56\% &  1 \\ \hline
 \bf{Our baseline}  & 93.70\% &  32 \\ \hline
 \bf{Our ternary}  & 93.66\% &  2 (ternary) \\ \hline
\bf{Our binary}  & 93.53\% &  1 \\ \hline 
\end{tabular}
\end{table}

\emph{\textbf{Binary Weight Quantization Results on ResNet for ImageNet Dataset}}: 
The binarization of ResNet models on ImageNet data set is widely acknowledged as a very challenging task. As a result, there are very limited prior work (e.g., the one-shot ADMM \cite{leng2017extremely}) with binarization results on ResNet models. As \cite{leng2017extremely} targets ResNet-18 (which is even more challenging than ResNet-50 or larger ones), we make a fair comparison on the same model. Table \ref{table:ResNet-quan} demonstrates the comparison results (Top-5 accuracy loss).
  In prior work, it is by default that the first and last layers are not quantified (or quantized to 8 bits) as these layers have a significant effect on overall accuracy. When leaving the first and last layers unquantized, our framework is not progressive, but an extended one-shot ADMM-based framework. We can observe the higher accuracy compared with the prior method under this circumstance (first and last layers unquantized while the rest of layers binarized).
The Top-1 accuracy has similar result: 3.8\% degradation in our extended one-shot and 4.3\% in \cite{leng2017extremely}.

\begin{table}[ht]
\centering
\caption{Comparisons of weight quantization results on ResNet-18 for ImageNet data set.}\label{table:ResNet-quan}
\begin{tabular}{p{2.7cm}p{2cm}p{2.2cm}}
\hline
Method & Relative Top-5 acc. loss  & Num. of bits \\ 
\hline
Uncompressed & 0.0\%  & 32 \\ \hline
 {One-shot ADMM quantization \cite{leng2017extremely}}  & 2.9\%  & 1 (32 for the first and last) \\ 
\hline
\bf{Our method (one-shot)}  & 2.5\%  & 1 (32 for the first and last)  \\ \hline
\bf{Our method}  & 5.8\%  & 1 \\ \hline
\end{tabular}
\end{table}

Using the progressive framework, we can derive a \underline{fully binarized ResNet-18}, in which weights in all layers are binarized. The accuracy degradation is 5.8\%, which is noticeable and shows that the full binarization of ResNet is a challenging task even under the progressive framework. We did not find prior work for comparison on this result.

\section{Conclusion}

In this work, we extended the prior one-shot ADMM-based framework and developed a multi-step, progressive DNN weight pruning and quantization framework, in which we achieve  further  weight  pruning/quantization and provide faster convergence rate. We achieve 246$\times$, 36$\times$, and 8$\times$ weight pruning on LeNet-5, AlexNet, and ResNet-50 models, respectively, with (almost) zero accuracy loss. We also derive the first lossless, fully binarized (for all layers) LeNet-5 for MNIST and VGG-16 for CIFAR-10.

%
%
%

\section*{Supplementary Materials}

\section*{Stability of Hyperparameter}

We test the stability of our hyperparameter on VGG-16 for CIFAR-10 data set.
Figure 6 demonstrates that our method is stable on parameter $\rho$ (the major hyperparameter in ADMM regularization). In our experiment, we change $\rho$ from 0.0005 to 0.005 with the same pruning rate, and the accuracy we achieve is close for different $\rho$ values.

\begin{figure} [!ht]
     \centering
     \includegraphics[width=\columnwidth]{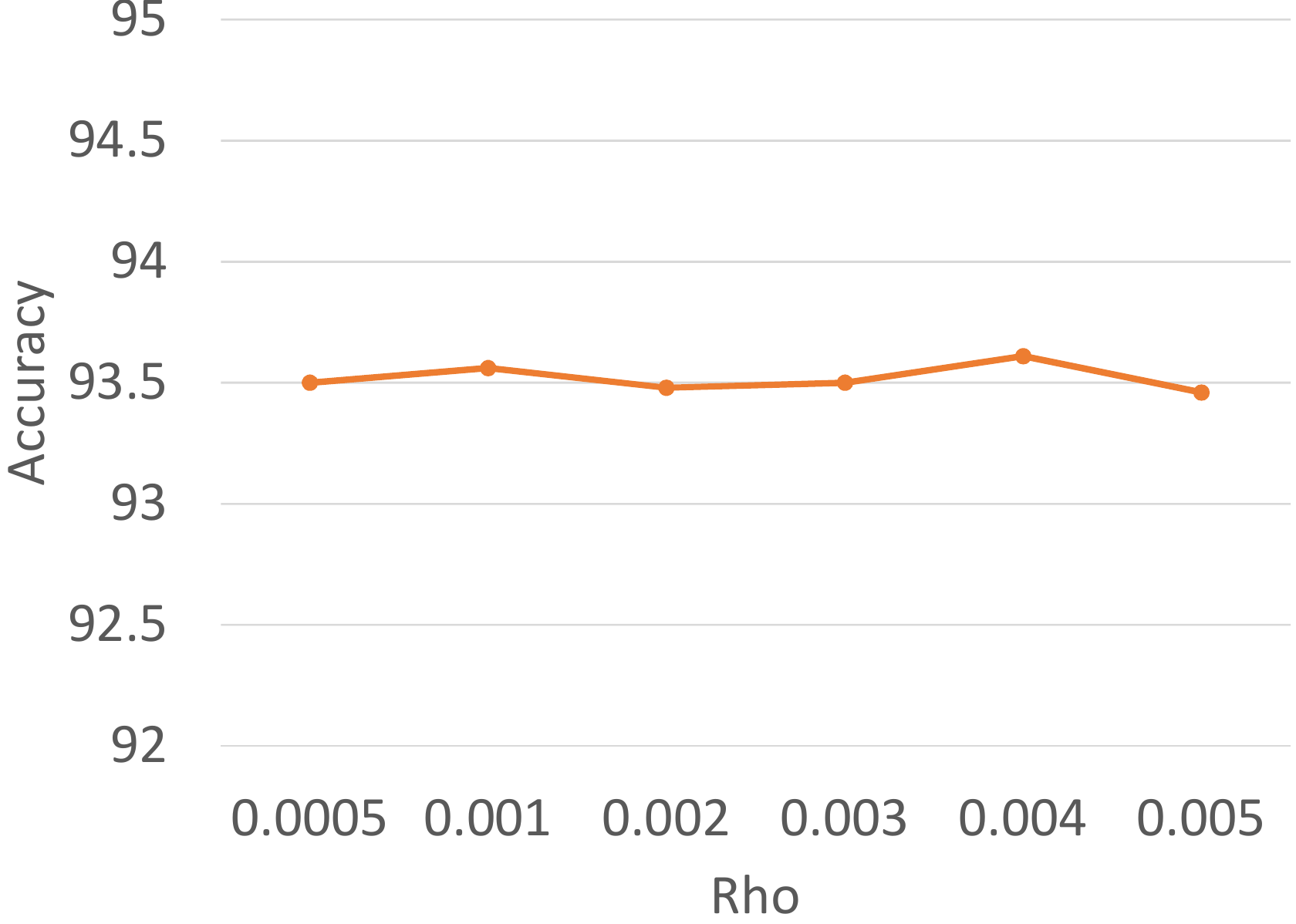}    
     \caption{Illustration of the stability of accuracy on $\rho$ values.}
     \label{fig1}
 \end{figure}

\section*{Convergence Analysis}

We test the convergence of our method on VGG-16 for CIFAR-10 data set.
Figure 7 demonstrates that our method (ADMM regularization) achieves fast convergence rate, where the gap between $W_{k+1}$ and $Z_{k+1}$ converges to zero in around 7 ADMM iterations.

\begin{figure} [!ht]
     \centering
     \includegraphics[width=\columnwidth]{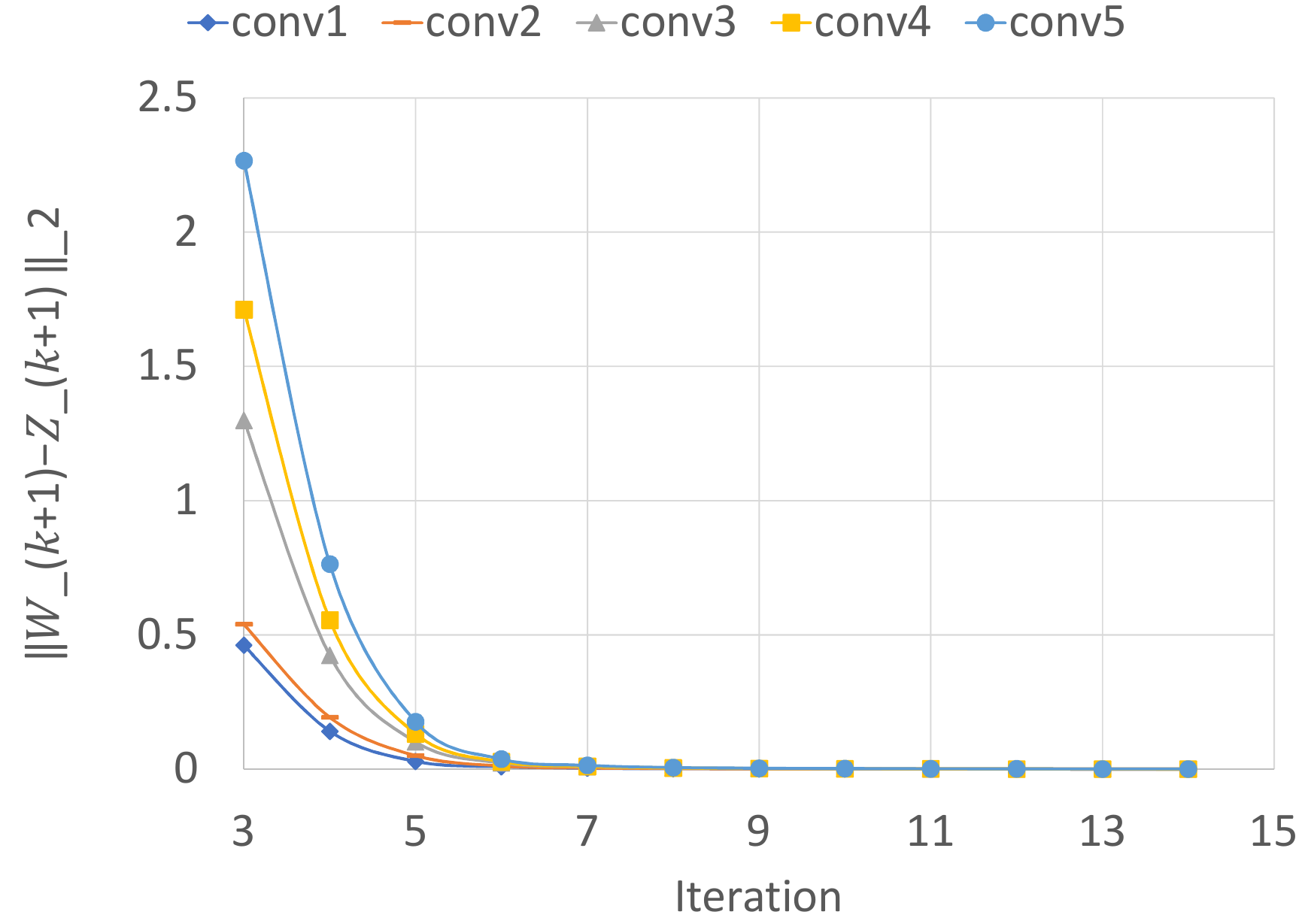}    
     \caption{Convergence results of ADMM regularization in our method.}
     \label{fig3}
 \end{figure}

\section*{Structured Weight Pruning Results}

Models of the following structured weight pruning results are in the anonymous link \url{https://bit.ly/2TYx7Za}. These results significantly outperform prior arts (if any). We focus on column pruning except for MobileNet V2, which is more suitable for filter pruning.

Table 8 shows our column pruning result on VGG-16 for CIFAR-10 data set. We achieve 29$\times$ structured pruning rate with 0.34$\%$ accuracy loss.

Table 9 shows our filter pruning result on MobileNet V2 for CIFAR-10 data set. We achieve 7.1$\times$ structured pruning rate (which is very difficult for MobileNet) with 0.2$\%$ accuracy loss.

Table 10 shows our column pruning result on LeNet-5 for MNIST data set. We achieve 37.1$\times$ structured pruning rate with 0.18$\%$ accuracy loss.

Table 11 shows our column pruning result on ResNet-18 for ImageNet data set. We achieve 3$\times$ structured pruning rate without any accuracy loss. The best in prior work results in at least 1\% accuracy loss with 2$\times$ structured pruning rate.

\begin{table}[!ht]
\centering
\caption{Column pruning results on VGG-16 for CIFAR-10 data set. }
\begin{tabular}{p{2cm}p{5cm}}
\hline
Method &     Prune rate / Top 1 accuracy loss  \\ \hline
baseline & 1$\times$ / 0\% \\ \hline
\bf{Our method}   &   9.3$\times$ / $-$0.06\% \\ \hline
\bf{Our method} & 29.0$\times$ / 0.34\% \\ \hline
\end{tabular}
\end{table}

\begin{table}[!ht]
\centering
\caption{Filter pruning results on Mobilenet V2 for CIFAR-10 data set. }
\begin{tabular}{p{2cm}p{5cm}}
\hline
Method &     Prune rate / Top 1 accuracy loss  \\ \hline
baseline & 1$\times$ / 0\% \\ \hline
\bf{Our method} & 7.1$\times$ / 0.20\% \\ \hline
\end{tabular}
\end{table}

\begin{table}[!ht]
\centering
\caption{Column pruning results on LeNet-5 for MNIST data set. }
\begin{tabular}{p{2cm}p{5cm}}
\hline
Method &     Prune rate / Top 1 accuracy loss  \\ \hline
baseline & 1$\times$ / 0\% \\ \hline
\bf{Our method}   &   17.7$\times$ / 0.05\% \\ \hline
\bf{Our method} & 37.1$\times$ / 0.18\% \\ \hline
\bf{Our method} & 105.5$\times$ / 0.87\% \\ \hline
\end{tabular}
\end{table}

\begin{table}[!ht]
\centering
\caption{Column pruning results on ResNet-18 for Imagenet data set. }
\begin{tabular}{p{2cm}p{5cm}}
\hline
Method &     Prune rate / Top 1 accuracy loss  \\ \hline
baseline & 1$\times$ / 0\% \\ \hline
\bf{Our method}   &   3.0$\times$ / 0.0\% \\ \hline

\end{tabular}
\end{table}

\section*{Combination of (Non-Structured) Weight Pruning and Quantization}

Models of the following results are released in the anonymous link \url{https://bit.ly/2TYx7Za}. We did not find prior work on the combination of ResNet non-structured weight pruning and quantization results.

\begin{table}[!ht]
\centering
\caption{Combination of nonstructured pruning and quantization on ResNet-50 for Imagenet data set. }
\begin{tabular}{p{2cm}p{5cm}p{2cm}}
\hline
Method &     Prune rate / Quantization bits & Accuracy loss  \\ \hline
baseline & 1$\times$ / 32 & 0\% \\ \hline
\bf{Our method}   &   8$\times$ / 6 & 0.2\% \\ \hline
\end{tabular}
\end{table}

\begin{table}[!ht]
\centering
\caption{Combination of nonstructured pruning and quantization on ResNet-18 for Imagenet data set. }
\begin{tabular}{p{2cm}p{5cm}p{2cm}}
\hline
Method &     Prune rate / Quantization bits & Accuracy loss  \\ \hline
baseline & 1$\times$ / 32 & 0\% \\ \hline
\bf{Our method}   &   5$\times$ / 5 & 0.0\% \\ \hline
\end{tabular}
\end{table}

Table 12 shows the combination of nonstructured pruning and quantization on ResNet-50 for Imagenet data set, in which we achieve 8$\times$ pruning rate and quantize the weights in 6 bits with 0.2\% (Top-5) accuracy loss, with baseline Top-5 accuracy 92.9\%.

Table 13 shows the combination of nonstructured pruning and quantization on ResNet-18 for Imagenet data set, in which we achieve 5$\times$ pruning rate and quantize the weights in 5 bits without accuracy loss (baseline Top-5 accuracy 89.1\%).

\end{document}